\documentclass{ieeetj}
\usepackage{cite}
\usepackage{amsmath,amssymb,amsfonts}
\usepackage{algorithmic}
\usepackage{graphicx,color}
\usepackage{textcomp}
\usepackage{xcolor}
\usepackage{hyperref}
\hypersetup{hidelinks=true}
\usepackage{algorithm}
\usepackage{algorithmic}
\usepackage[utf8]{inputenc}
\usepackage{tcolorbox}
\usepackage{multirow}
\usepackage{makecell}
\usepackage{setspace}
\usepackage{caption}

\def\BibTeX{{\rm B\kern-.05em{\sc i\kern-.025em b}\kern-.08em
    T\kern-.1667em\lower.7ex\hbox{E}\kern-.125emX}}

\begin{document}


\title{Large Language Model Enhanced Particle Swarm Optimization for Hyperparameter Tuning for Deep Learning Models}

\author{Saad Hameed\authorrefmark{1}, Basheer Qolomany\authorrefmark{2}, Member, IEEE, Samir Brahim Belhaouari\authorrefmark{1},\\ Senior Member, IEEE, Mohamed Abdallah\authorrefmark{1}, Senior Member, IEEE, Junaid Qadir\authorrefmark{3}, \\ Senior Member, IEEE, and Ala Al-Fuqaha\authorrefmark{1}, Senior Member, IEEE}
\affil{\authorrefmark{1}Division of Information and Computing Technology, College of Science and Engineering, Hamad Bin Khalifa University, Doha, 5825 Qatar}
\affil{\authorrefmark{2}Department of Medicine, College of Medicine, Howard University, Washington, D.C. 20059 USA}
\affil{\authorrefmark{3}Department of Computer Science and Engineering, Qatar University, Doha, 5825 Qatar}
\corresp{Corresponding author: Saad Hameed (email: saha41211@hbku.edu.qa).}

\begin{abstract}

Determining the ideal architecture for deep learning models, such as the number of layers and neurons, is a difficult and resource-intensive process that frequently relies on human tuning or computationally costly optimization approaches. While Particle Swarm Optimization (PSO) and Large Language Models (LLMs) have been individually applied in optimization and deep learning, their combined use for enhancing convergence in numerical optimization tasks remains underexplored. Our work addresses this gap by integrating LLMs into PSO to reduce model evaluations and improve convergence for deep learning hyperparameter tuning. The proposed LLM-enhanced PSO method addresses the difficulties of efficiency and convergence by using LLMs (particularly ChatGPT-3.5 and Llama3) to improve PSO performance, allowing for faster achievement of target objectives. Our method speeds up search space exploration by substituting underperforming particle placements with best suggestions offered by LLMs. Comprehensive experiments across three scenarios---(1) optimizing the Rastrigin function, (2) using Long Short-Term Memory (LSTM) networks for time series regression, and (3) using Convolutional Neural Networks (CNNs) for material classification---show that the method significantly improves convergence rates and lowers computational costs. Depending on the application, computational complexity is lowered by 20\% to 60\% compared to traditional PSO methods. Llama3 achieved a 20\% to 40\% reduction in model calls for regression tasks, whereas ChatGPT-3.5 reduced model calls by 60\% for both regression and classification tasks, all while preserving accuracy and error rates. This groundbreaking methodology offers a very efficient and effective solution for optimizing deep learning models, leading to substantial computational performance improvements across a wide range of applications.

\end{abstract}

\begin{IEEEkeywords}
Deep Learning Optimization, PSO, LLM, Machine Learning, Hyper-parameter Optimization.
\end{IEEEkeywords}


\maketitle

\section{INTRODUCTION}
Deep learning (DL), a subset of machine learning (ML), utilizes multi-layer neural networks to learn from data and automate model construction. This complexity enables DL to uncover intricate patterns in large datasets, often outperforming traditional methods in areas like image recognition, natural language processing, and autonomous systems. However, training these models effectively requires advanced techniques and significant computational resources.

A critical aspect of DL is optimizing hyper-parameters such as the number of layers and neurons, which significantly affect model performance. Manually tuning these parameters is time-consuming and can yield unreliable results. Grid search \cite{shekar2019grid}, \cite{belete2022grid} is commonly used, testing different combinations of hyper-parameters and selecting the best-performing model. However, this approach is computationally expensive, especially with increasing parameter precision, and may lead to suboptimal results if the chosen parameters are inadequate \cite{ganjisaffar2011distributed}.

\subsection{Metaheuristic Optimization Algorithms}

Optimization techniques aim to find optimal values for design variables by balancing objectives and constraints. Gradient-based methods \cite{haji2021comparison} efficiently navigate smooth, continuous problems but struggle with complex, noisy, or discrete scenarios due to their susceptibility to local optima.

Metaheuristic algorithms like GA \cite{hamdia2021efficient} and PSO \cite{shami2022particle} explore the entire search space to identify global optima \cite{mehboob2016genetic}. These problem-independent approaches, inspired by natural processes, are effective for multimodal and complex optimization tasks.

\textbf{Particle Swarm Optimization (PSO):} PSO, introduced by Kennedy and Eberhart \cite{kennedy1995particle}, leverages swarm intelligence to balance exploration and exploitation. With minimal parameters and computational simplicity, PSO is adaptable and efficient for diverse optimization challenges. Though prone to premature stagnation, variants involving parameter tuning and hybrid strategies address this limitation effectively.

PSO outperforms GA in continuous optimization \cite{adetunji2020review}, avoiding fitness sorting and scaling linearly with population size. Studies \cite{katiyar2010comparative} show PSO's computational efficiency and ability to deliver high-quality solutions with fewer evaluations, affirming its suitability for real-world applications. DE, while excelling in diversification, complements PSO's strengths \cite{kachitvichyanukul2012comparison}.

\subsection{Large Language Model Advances}

Large Language Models (LLMs) have gained prominence due to advancements in transformer architecture and self-supervised learning, enabling diverse applications across academia and industry. These models excel in addressing complex tasks in healthcare, environmental science, and beyond, while prompting discussions on their societal implications.

Recent LLM advancements, such as ChatGPT, Llama2, Llama3, and Phi-2, have achieved remarkable results in translation, question answering, and other tasks. Leveraging transformer architectures and extensive pretraining on vast datasets, these models support zero-shot and few-shot learning via prompt engineering. Fine-tuning further enhances their adaptability to specialized tasks, with techniques like in-context learning demonstrating improved performance through prompt-provided examples.

LLMs such as GPT and BERT exhibit strengths in translation, summarization, and other tasks, driving innovation in research and industry. Transformer architectures support versatile designs: encoder-only (e.g., BERT for text understanding), decoder-only (e.g., GPT for text generation), and encoder-decoder (e.g., T5 for translation and summarization), catering to varied application needs.

\subsection{Leveraging LLMs for Optimizing PSO}

Our novel approach integrates Large Language Models (LLMs) like ChatGPT-3.5 and Llama3 with Particle Swarm Optimization (PSO) to enhance optimization efficiency, referred to as LLM-driven PSO. This hybrid method leverages LLMs to suggest optimal particle positions, enabling faster PSO convergence on various objective functions. The goal is to showcase the contribution of LLMs within a straightforward PSO framework.

We evaluate LLM-driven PSO in three scenarios: (1) optimizing the Rastrigin Function, a benchmark for PSO \cite{redoloza2021comparison}, \cite{bansal2011inertia}; (2) optimizing layers and neurons in an LSTM model for time series prediction; and (3) optimizing layers and filters in a CNN for classifying materials as recyclable or organic. Standard PSO was initially applied to these cases, observing convergence rates while dynamically tuning DL model hyperparameters. The LLM-driven PSO approach demonstrated superior performance by achieving desired results more efficiently and reducing DL model evaluations. Results using ChatGPT-3.5 and Llama3 are discussed in detail.

\subsection{Paper's Contributions and Organization}

The key contributions of the paper are: 

\begin{itemize}
    \item We propose a novel LLM-driven PSO framework that integrates Large Language Models to enhance the convergence speed of optimization algorithms. We validate its utility across three distinct applications: (1) optimizing the Rastrigin function as a standard PSO benchmark, (2) improving the performance of an LSTM-based regression model for air quality index prediction, and (3) advancing a CNN-based classification model for distinguishing recyclable from organic materials.

    \item We provide a detailed analysis of how the exploration and exploitation dynamics in PSO can be strategically modified using LLM-driven insights to accelerate convergence. The findings establish the advantages of our approach compared to traditional PSO methods.  

    \item We introduce an optimization strategy that leverages selective LLM guidance to reduce the number of PSO iterations while maintaining performance. This hybrid approach enhances the efficiency of metaheuristic methods and offers a systematic framework for combining LLMs with PSO.  

    \item We demonstrate the computational efficiency of our method by significantly reducing the number of LLM invocations required for achieving target accuracy. This efficiency is quantified with a reduction of 20\%–40\% for Llama3 and 60\% for ChatGPT-3.5 in regression tasks, and a 60\% reduction for both models in classification tasks. These results highlight the practical benefits of our approach in terms of resource optimization and computational cost savings.  
    
\end{itemize}

The rest of the paper is organized as follows. Section \ref{sec:related_work} goes through the related work. Section \ref{sec:methodology} outlines the proposed methodology, while Section \ref{sec:experimental_setup} provides a detailed description of the experimental setup. Section \ref{sec:results} reports the conducted experiments and the experimental results. Section \ref{sec:lessons_learned} summarizes the work and discusses the lessons learned during this research. Finally, Section \ref{sec:conclusion_and_future_work} concludes the work and states the future work.


\section{RELATED WORK}
\label{sec:related_work}

\subsection{LLMs in Natural Language Processing (NLP)}

Recent advancements in NLP have shifted towards task-agnostic pre-training, leading to significant performance improvements. For example, \cite{radford2019language} showed that a single pre-trained language model could perform zero-shot tasks, although with lower accuracy than supervised models. Building on this, \cite{brown2020language} scaled the model to 175 billion parameters (GPT-3), achieving superior performance in zero-shot, one-shot, and few-shot tasks. Notably, GPT-3's few-shot performance was competitive with, and sometimes surpassed, state-of-the-art fine-tuned models, showcasing the enhanced meta-learning capabilities of larger models.

\subsection{PSO-Based Optimization in Machine Learning}

PSO has been effectively applied to optimize ML models. For instance, \cite{fouad2021hyper} used PSO to automate CNN hyperparameter selection for digit recognition, achieving competitive results. \cite{utama2022pso} optimized CNN architecture for time-series data, outperforming traditional methods in hyperparameter tuning.

In SVM optimization, \cite{kalita2020svm} applied multi-PSO to improve classification accuracy in network intrusion detection, while \cite{tayebi2022deep} used PSO to optimize DNNs for fraud detection, outperforming grid search. Similarly, \cite{qolomany2017parameters} demonstrated PSO's efficiency in tuning DL model parameters, reducing search time significantly.

Qolomany et al. \cite{9322464} proposed a PSO-based approach for optimizing hyperparameters in Federated Learning, reducing communication rounds while maintaining accuracy. This method was effective in minimizing computational costs in smart city and industrial IoT case studies. 

In a more recent work, \cite{parkavi2024enhancing} proposed a decision support system combining Chaotic Particle Swarm Optimization (C-PSO) with explainable AI techniques to assess student performance. Their model demonstrated strong predictive performance and outperformed several metaheuristics, highlighting the utility of advanced PSO variants in optimizing real-world machine learning tasks. Salem et al. \cite{salem2024particle} recently employed PSO for optimizing machine learning hyperparameters, significantly improving predictive performance in complex epidemiological data contexts such as COVID-19 data analysis. Pranolo et al. \cite{pranolo2023optimized} utilized PSO to optimize architectures for LSTM, CNN, and MLP models, highlighting the superiority of PSO-tuned deep learning models for accurate time-series forecasting of air quality.

\subsection{Optimization Using Large Language Models (LLMs)}

Optimization is vital in many fields, but gradient-based methods struggle when gradients are unavailable. Recent approaches leverage LLMs for optimization. For instance, OPRO \cite{yang2024largelanguagemodelsoptimizers} uses LLMs to generate solutions from previous results and associated values, showing promise in tasks like linear regression and the traveling salesman problem.

\cite{ma2024large} advanced this by exploring Automatic Prompt Optimization but highlighted challenges such as inaccurate reflection of errors and ineffective prompts. They introduced "Automatic Behavior Optimization" to address these issues and improve control over model behavior.

Other works have focused on specific tasks, such as APE \cite{zhou2022large}, which uses LLMs to generate and optimize instructions, and MyCrunchGPT \cite{kumar2023mycrunchgpt}, which leverages ChatGPT to optimize Scientific Machine Learning workflows. In hyperparameter optimization, \cite{liu2024large} introduced AgentHPO, utilizing two agents to optimize parameters efficiently. \cite{zhang2023using} demonstrated that LLMs can outperform traditional methods like random search and Bayesian optimization in hyperparameter tuning with smaller search budgets. 

\begin{figure*}[!t]
  \centering
  \includegraphics[scale=1]{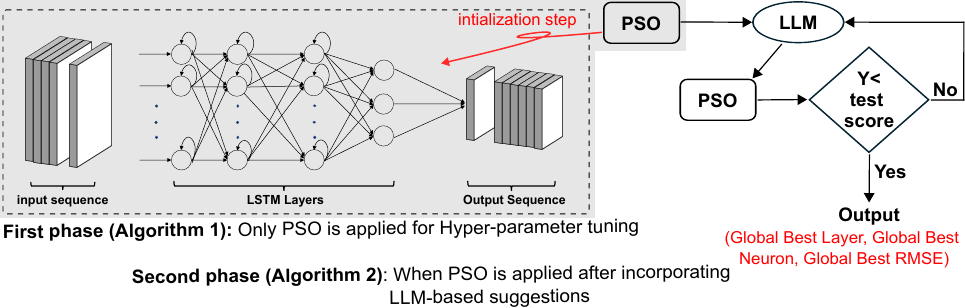} 
  \caption{Two algorithms are illustrated. \textbf{First phase (Algorithm 1):} Standard PSO optimizes DL model layers and neurons (shaded area). \textbf{Second phase (Algorithm 2):} Key variables—\textit{Global Best Layer}, \textit{Global Best Neuron}, and \textit{Global Best RMSE}—are computed. After initial PSO-based DL model setup, LLM suggests improved particle positions and velocities for faster convergence. \textit{`test\_score'} (from Algorithm 1) and \textit{`y'} (from Algorithm 2) are compared to decide whether to query the LLM again.}

  \label{methodology_diagram}
\end{figure*}

\subsection{Our Contribution: LLM-Driven PSO for DL Optimization}
To the best of our knowledge, the PSO algorithm has not been previously combined with an LLM for optimizing ML algorithms. In this study, we introduce a novel method, LLM-driven PSO, which dynamically updates the hyperparameters of DL models, reducing the frequency of model calls and decreasing computational overhead. Unlike previous work that focuses on improving accuracy through iterative optimization, our approach specifically addresses the challenges in optimizing deep learning models, such as determining the optimal network topology, which is both labor-intensive and computationally demanding. Mumtahina et al. \cite{mumtahina2024hyperparameter} provided a recent systematic review highlighting the effectiveness of metaheuristic optimization algorithms, including PSO, for tuning hyperparameters, underscoring the relevance and timeliness of our contribution in integrating LLM with PSO to enhance deep learning model optimization.


\section{Methodology}
\label{sec:methodology}

Our research focuses on optimizing DL models using PSO enhanced by an LLM—specifically OpenAI’s ChatGPT-3.5 and Meta’s Llama3. We evaluate the approach through three case studies, targeting key hyperparameters such as layers, neurons, and filters. The methodology consists of two phases. In the \textit{first phase (Algorithm 1)}, standard PSO is applied independently to optimize the DL model, recording the number of iterations or model calls required to reach the objective function (\textit{test\_score}). In the \textit{second phase (Algorithm 2)}, we integrate LLM suggestions into the PSO process: after a few PSO iterations, the LLM recommends improved particle positions and velocities to accelerate convergence. Our goal is to demonstrate the contribution of LLMs within a simple PSO setup. Figure~\ref{methodology_diagram} illustrates this process, with the \textit{first phase} shown in the shaded area and the \textit{second phase} covering both shaded and unshaded regions. 

In the \textit{second phase (Algorithm 2)}, PSO begins with a fixed number of iterations using randomly initialized particles (\textit{initialization step}). LLM-generated particle positions and velocities are then incorporated into a second PSO block to enhance convergence. The worst-performing particles from the initial phase are replaced by the best LLM suggestions, yielding an updated model with revised layers and neurons and a new output \textit{y}. This output is compared with the \textit{test\_score} from the \textit{first phase (Algorithm 1)}. If \textit{y} meets the objective, it is accepted; otherwise, additional PSO iterations are performed.

We define three variables to track key parameters during the \textit{first phase (Algorithm 1)} and at the end of the \textit{second phase (Algorithm 2)}. \textit{Global\_Best\_Layer} records the optimal number of layers from the best particle positions after convergence. \textit{Global\_Best\_Neuron} captures the optimal number of neurons based on the best particle velocities. \textit{Global\_Best\_Cost} stores the final cost achieved using the best combination of layers and neurons/filters.

We conduct experiments across three distinct cases, each with a different objective function and optimized value. First, we use the Rastrigin function—a standard PSO benchmark—with a global minimum of 0. The Rastrigin function, a standard benchmark with high complexity and many local minima, was chosen to rigorously evaluate the effectiveness of PSO and our LLM-enhanced variant. Second, we evaluate our approach on a regression task using an LSTM model to predict next-day Air Quality Index (AQI). Finally, we apply it to a classification task using a CNN to distinguish between recyclable and organic materials. The final cost is measured by RMSE for regression and accuracy for classification.

\subsection{\textbf{Particle Swarm Optimization (PSO)}}

In PSO, random particles are generated in certain positions and with certain velocities. These positions and velocities are updated by the following mathematical expressions. 









{%
\setlength{\abovedisplayskip}{3pt}
\setlength{\belowdisplayskip}{3pt}

\textbf{Particle velocity update:}

\begin{equation}
\begin{split}
    Vl_{id}(t+1) &= w \cdot Vl_{id}(t) \\
    &\quad + c_1 \cdot r_1 \cdot (pbest_{id} - Xl_{id}(t)) \\
    &\quad + c_2 \cdot r_2 \cdot (gbest_d - Xl_{id}(t))
\end{split}
\end{equation}

\begin{equation}
\begin{split}
    Vn_{id}(t+1) &= w \cdot Vn_{id}(t) \\
    &\quad + c_1 \cdot r_1 \cdot (pbest_{id} - Xn_{id}(t)) \\
    &\quad + c_2 \cdot r_2 \cdot (gbest_d - Xn_{id}(t))
\end{split}
\end{equation}

\textbf{Particle position update:}

\begin{equation}
    Xl_{id}(t+1) = Xl_{id}(t) + Vl_{id}(t+1)
\end{equation}

\begin{equation}
    Xn_{id}(t+1) = Xn_{id}(t) + Vn_{id}(t+1)
\end{equation}
}


\begin{table}[!h]
\caption{Parameters used in our experiments}
\centering
\renewcommand{\arraystretch}{1.05} 
\scriptsize 
\begin{tabular}{|p{3cm}|p{4.2cm}|}
\hline
\textbf{Parameter} & \textbf{Value} \\
\hline
Population size & 5, 10, 15, 20, 50, 100 \\
\hline
$c_1$, $c_2$ (exploration/exploitation) & Uniformly sampled from [0, 1] \\
\hline
Max. iterations & 10, 20, 30, 40, 50 \\
\hline
Hidden layers & Range: [2, 5] \\
\hline
Neurons per layer & Range: [2, 200] \\
\hline
Particle dimensions & Hidden layers and neurons per layer \\
\hline
Layer velocity & Min = 1, Max = 0.2(MaxLayers $-$ MinLayers) \\
\hline
Neuron velocity & Min = 1, Max = 0.2(MaxNeuron $-$ MinNeuron) \\
\hline
\end{tabular}
\label{PSO_parameters}
\end{table}

Where:
\begin{itemize}
    \item $Vl_{id}(t)$ are the layers velocity of particle $i$ in dimension $d$ at iteration $t$,
    \item $Vn_{id}(t)$ are the neurons velocity of particle $i$ in dimension $d$ at iteration $t$,
    \item $w$ is the inertia weight controlling the impact of the previous velocity,
    \item $c_1$ and $c_2$ are acceleration coefficients (exploration and exploitation components, respectively),
    \item $r_1$ and $r_2$ are random values sampled from a uniform distribution [0, 1],
    \item $pbest_{id}$ is the personal best position of particle $i$ in dimension $d$,
    \item $Xl_{id}(t)$ are the layers position of particle $i$ in dimension $d$ at iteration $t$,
    \item $Xn_{id}(t)$ are the neurons position of particle $i$ in dimension $d$ at iteration $t$,
    \item $gbest_d$ is the global best position in dimension $d$.
\end{itemize}

\begin{algorithm}
\scriptsize 
\caption{DL Model Optimization using PSO}
\label{alg:PSO_DLOptimization}
\begin{algorithmic}[1]
\STATE \textbf{Input:} PM2.5 pollutant values (regression) and material images (classification), time\_stamp, $c_1$, $c_2$, inertia weight $W$, population size (PopSize), max iterations (MaxIt), range bounds: MinLayer, MaxLayer, MinNeurons, MaxNeurons, MaxLayerVelocity, MaxNeuronVelocity.
\STATE \textbf{Output:} Optimal configuration of hidden layers $L$ ($X_l$, $V_l$) and neurons $N$ ($X_n$, $V_n$).
\STATE \textbf{Begin:}
\STATE Initialize $c_1$, $c_2$, $W$, PopSize, MaxIt, range bounds. Define objective function (RMSE for regression, accuracy for classification). Randomly generate particles (L, N); evaluate fitness; assign personal best $p_{best}$ and global best $g_{best}$.
\STATE \textbf{While} stopping criterion not met:
\STATE \hspace{0.5em} Update L, N, and velocities using Equations (1)–(4); evaluate fitness;
\STATE \hspace{0.5em} Update $p_{best}$ if improved; update $g_{best}$ if surpassed.
\STATE \textbf{Return:} Optimal L and N.
\STATE \textbf{End}
\end{algorithmic}
\end{algorithm}



Table \ref{PSO_parameters} summarizes the configuration of our PSO algorithm setup. Algorithm 1 outlines our proposed PSO-based parameter selection method for DL models.

\subsection{\textbf{Large Language Model (ChatGPT \& Llama3)}}

Large language models \cite{kasneci2023chatgpt} are transforming AI with applications in content creation, engagement, and personalized learning. Their integration demands critical thinking, fact-checking, and addressing challenges like bias to ensure ethical use.

\textbf{ChatGPT:}
ChatGPT, developed by OpenAI, has transformed human–AI interaction \cite{kalla2023study}. GPT-4, released in March 2023, introduced multimodal capabilities like image captioning and chart reasoning \cite{wu2023brief}. Its evolution from GPT-1 to GPT-4 integrates deep learning, instruction tuning, in-context learning, and reinforcement learning.

\textbf{Llama3:}
Llama3 \cite{dubey2024llama}, Meta’s latest foundation model series, includes up to 405 billion parameters and a 128K token context window, excelling in multilingualism, reasoning, and tool use. With 50× more FLOPs than Llama2, it delivers state-of-the-art performance \cite{huang2024good}. Llama3 is open-source and being extended for multimodal tasks in image, video, and speech.

\begin{algorithm}
\scriptsize 
\caption{DL Model Optimization using LLM-driven PSO}
\begin{algorithmic}[1]
\STATE \textbf{Input:} Prompt with particle positions ($X_l$, $X_n$) and velocities ($V_l$, $V_n$).
\STATE \textbf{Output:} Updated particle positions favoring faster convergence.
\STATE \textbf{Begin:}
\STATE Perform Steps 4 and 5 from Algorithm~1.
\STATE Query LLM for new $X_l$ and $X_n$ suggestions.
\STATE Evaluate fitness for suggested positions; identify best cost particle.
\STATE Replace worst PSO particles with LLM-suggested best.
\STATE Repeat Steps 3–6 until $g_{best}$ is unchanged or MaxIt is reached.
\STATE \textbf{Return:} Optimal configuration of hidden layers ($L$) and neurons ($N$).
\STATE \textbf{End}
\end{algorithmic}
\end{algorithm}
\begin{figure*}[!h]
\scriptsize 
\begin{tcolorbox}[colback=white, colframe=black, boxrule=0.2mm, arc=0mm, title=Format of our LLM Prompt, label=box:llm_prompt]

\textcolor{blue}{\textit{Npop = 5, 10, 15, 20, 50, or 100}}\\
\textcolor{blue}{\textit{particle\_prompt\_string = 80, 3, 1.6, 1.2, 0.1342, 120, 4, 1.8, 1.5, 0.1030, 95, 2, 1.6, 1, 0.0012, .......}}

\textcolor{red}{\texttt{\textcolor{black}{my\_prompt =} Below is the string showing the best number of neurons as the first entry and best number of layers as the second entry of the DL model for \textcolor{blue}{\{Npop\}} particles with their corresponding cost as the fifth entry, while dynamically updating the number of neurons and layers to reduce the cost for the same model using Particle Swarm Optimization. The third and the fourth entries are the neurons velocities and layers velocities, respectively. The first entry (Neurons) of the string ranges from 2 to 200, while the second entry (Layers) of the string ranges from 2 to 5.}}

\textcolor{blue}{\texttt{\{particle\_prompt\_string\}}}

\textcolor{red}{\texttt{Give me exactly \textcolor{blue}{\{Npop\}} more number of neurons and layers for the same model in order to reduce the cost further. Your response must be exactly in the same format as input and must contain only values. Your response must not contain the cost values.}}

\end{tcolorbox}
\end{figure*}

\subsection{\textbf{LLM-Driven PSO (Our Proposed Technique)}}

In our study, we integrate ChatGPT-3.5 and Llama3 into the PSO process using a prompt-based mechanism. The LLMs receive particle positions and velocities generated by PSO and return an equal number of improved, distinct suggestions to accelerate convergence. \textit{Algorithm 2} outlines this LLM-PSO parameter selection method. Initially, \textit{Algorithm 1} runs standalone PSO; from step 3 onward, the LLM is prompted for optimized positions and velocities. If the LLM's suggestions yield better fitness values, they replace the worst-performing PSO particles, updating the local and global bests that determine the optimal number of layers and neurons.

\section{Experimental Setup}
\label{sec:experimental_setup}
The finalized prompt for ChatGPT-3.5 and Llama3 is shown in the rectangular window. After iterative testing, we designed a prompt that reliably outputs a format compatible with our code. It defines variables such as \textit{Npop} (population size) and \textit{particle\_prompt\_string}, which includes particle positions, velocities, and cost values. Each particle's position and velocity are two-dimensional, representing the number of layers and neurons.

\subsection{\textbf{Rastrigin Function}}

The Rastrigin function is a non-convex, multimodal benchmark widely used to evaluate global optimization algorithms. Defined for an \( n \)-dimensional space, it is given by:

\begin{equation}
    f(x) = An + \sum_{i=1}^n \left[ x_i^2 - A \cos(2\pi x_i) \right]
\end{equation}

where \( A = 10 \), \( n = 2 \), and \( x_i \in [-5.12, 5.12] \). The global minimum is at \( x = 0 \) with \( f(x) = 0 \), while numerous local minima—especially near \( x_i \approx \pm 4.523 \)—pose challenges for local optimization.

To assess PSO performance, we tested particle counts of 20, 50, and 100 under three configurations: balanced (\( c_1 = c_2 = 0.5 \)), exploration-dominant (\( c_1 = 0.8, c_2 = 0.2 \)), and exploitation-dominant (\( c_1 = 0.2, c_2 = 0.8 \)). Each setup was run 10 times, recording the mean (\( \mu \)) and standard deviation (\( \sigma \)) of convergence iterations. Results are shown in Table \ref{Exploration_Exploitation_variations}.

Next, we implemented our LLM-driven PSO using a single configuration with \( c_1 = c_2 \). PSO runs for a few iterations, after which ChatGPT-3.5 or Llama3 provides optimization guidance. PSO then resumes using the LLM's suggestions, with further LLM calls made dynamically as needed.

We evaluated this scheme using the same particle counts as in the first phase (20, 50, and 100 particles) and recorded the results in Table \ref{Rastrigin_LLM_results}.
\begin{figure*}[]
  \centering
  \includegraphics[scale=0.30]{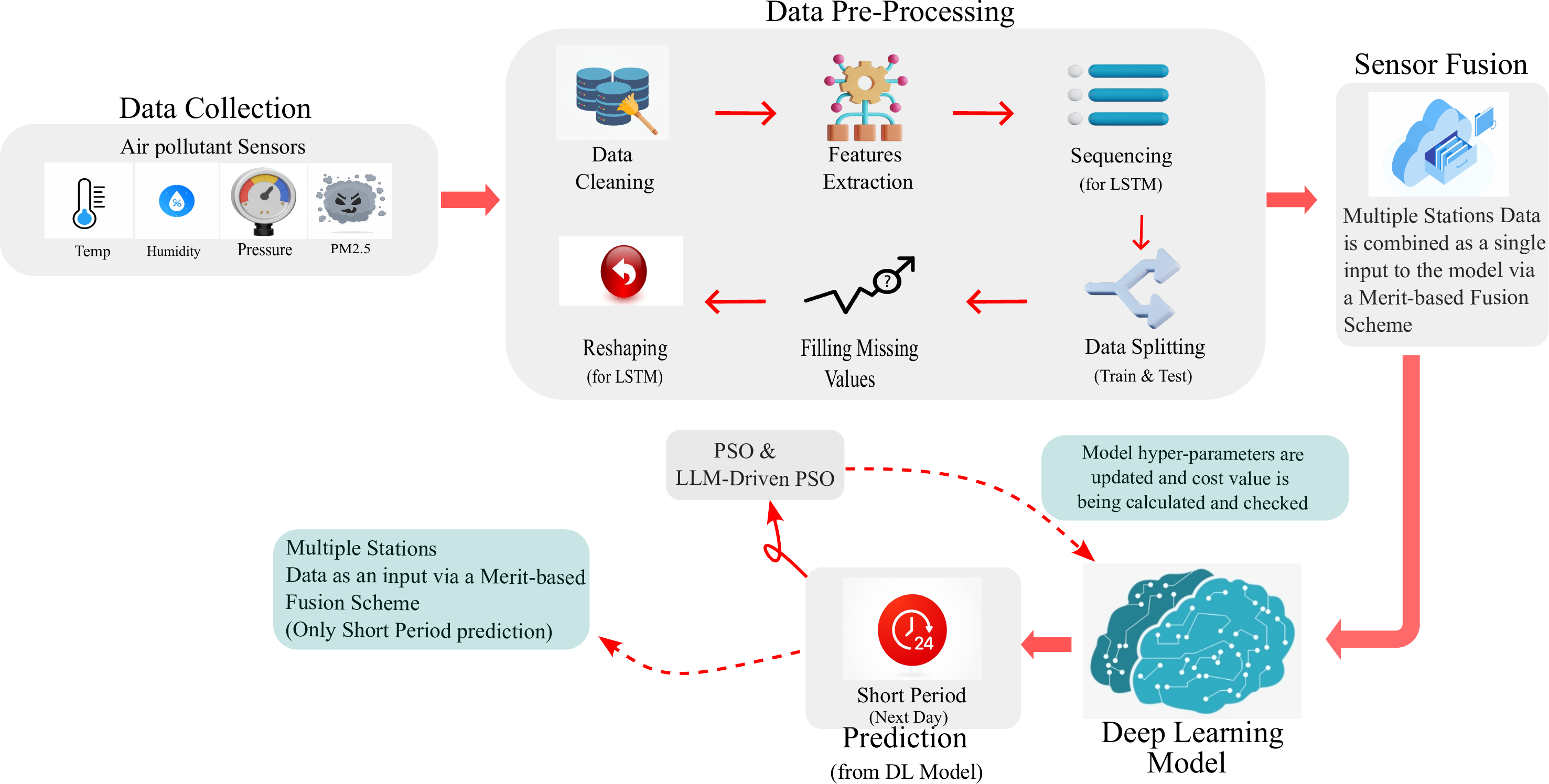} 
  \caption{Flow chart of the experimental setup, including data collection~\cite{hameed2023deep}, preprocessing, fusion, DL model training, and next-day prediction. PSO or LLM-driven PSO is used for hyperparameter optimization once the desired error is reached.
}
  \label{flow_chart}
\end{figure*}
\subsection{\textbf{Regression Model}}

LSTM models are well-suited for time-series analysis, and our study focuses on their application for AQI prediction, specifically building on prior work using sensor and weather data~\cite{hameed2023deep}. Our method improves LSTM hyperparameter tuning by reducing computational overhead and enhancing performance, which is critical for complex models where efficiency and predictive accuracy are tightly coupled.

The LSTM architecture comprises four components: the input, forget, and output gates, along with the cell state. These gates regulate what information is added, discarded, or passed on, while the cell state integrates prior memory and new input, as defined in Equations~\ref{input_gate}–\ref{cell_state}.

Additionally, ${\sim}{c}_t$ represents the intermediate cell state, computed using Equation \ref{cell_bar_gate}, and the hidden state $h_t$ encapsulates the block's output information (Equation \ref{hidden_state}). Here, $W_i$, $W_f$, and $W_o$ denote weight matrices for the input, forget, and output gates, respectively. The ReLU activation function, $f(x) = \max(0, x)$, ensures non-negative outputs by setting negative inputs to zero while preserving positive values.







{%
\setlength{\abovedisplayskip}{3pt} 
\setlength{\belowdisplayskip}{3pt} 
\begin{equation}
    i_t = \sigma(W_i.[h_{t-1}, x_t] + b_i)
    \label{input_gate}
\end{equation}

\begin{equation}
    f_t = \sigma(W_f.[h_{t-1}, x_t] + b_f)
    \label{forget_gate}
\end{equation}

\begin{equation}
    o_t = \sigma(W_o.[h_{t-1}, x_t] + b_o)
    \label{output_gate}
\end{equation}

\begin{equation}
    c_t = f_t \odot c_{t-1} + i_t \odot {\sim}{c}_t 
    \label{cell_state}
\end{equation}

\begin{equation}
    {\sim}{c}_t = ReLU(W_c.[h_{t-1}, x_t] + b_c)
    \label{cell_bar_gate}
\end{equation}

\begin{equation}
    h_t = o_t  \odot ReLU(c_t)
    \label{hidden_state}
\end{equation}
}%

The study in \cite{hameed2023deep} proposed a multimodal AQI prediction model combining LSTM for temporal patterns and YOLOv8 for vehicle detection using CCTV data. Real-time inputs from 10 sensors, 3 weather stations, and 16 CCTV cameras in Dalat City, Vietnam, were used to address challenges such as sensor outages and noisy data. Sensor fusion enabled accurate AQI forecasting, even with incomplete inputs.

In our study, we used multiple sensor inputs to predict AQI in Dalat City, following the six phases shown in Figure \ref{flow_chart}:

\begin{itemize}
    \item Collection of real-time data from CCTV, sensors, and weather stations.
    \item Pre-processing to ensure consistency and quality.
    \item Sensor fusion to combine multimodal inputs.
    \item Training an LSTM model on the fused data.
    \item Short-term AQI prediction using the trained model.
    \item Evaluation and parameter tuning based on optimization results.
\end{itemize}

Key challenges included data inconsistency and limited datasets. We trained on 18 weeks of data and tested on 3. PSO was used to optimize the LSTM's layers and neurons with 5 particles over 10 runs, totaling 50 model calls to reach the minimum cost.

We then applied our LLM-driven PSO using the same setup. PSO initially adjusted particle positions and velocities to optimize the LSTM model. After a few iterations, LLMs suggested replacements for the worst-performing particles, refining the model until the cost matched or surpassed that of PSO alone. The full process, including layer and neuron updates, is shown in Figure \ref{methodology_diagram}.

\begin{figure}
  \centering
  \includegraphics[scale=0.30]{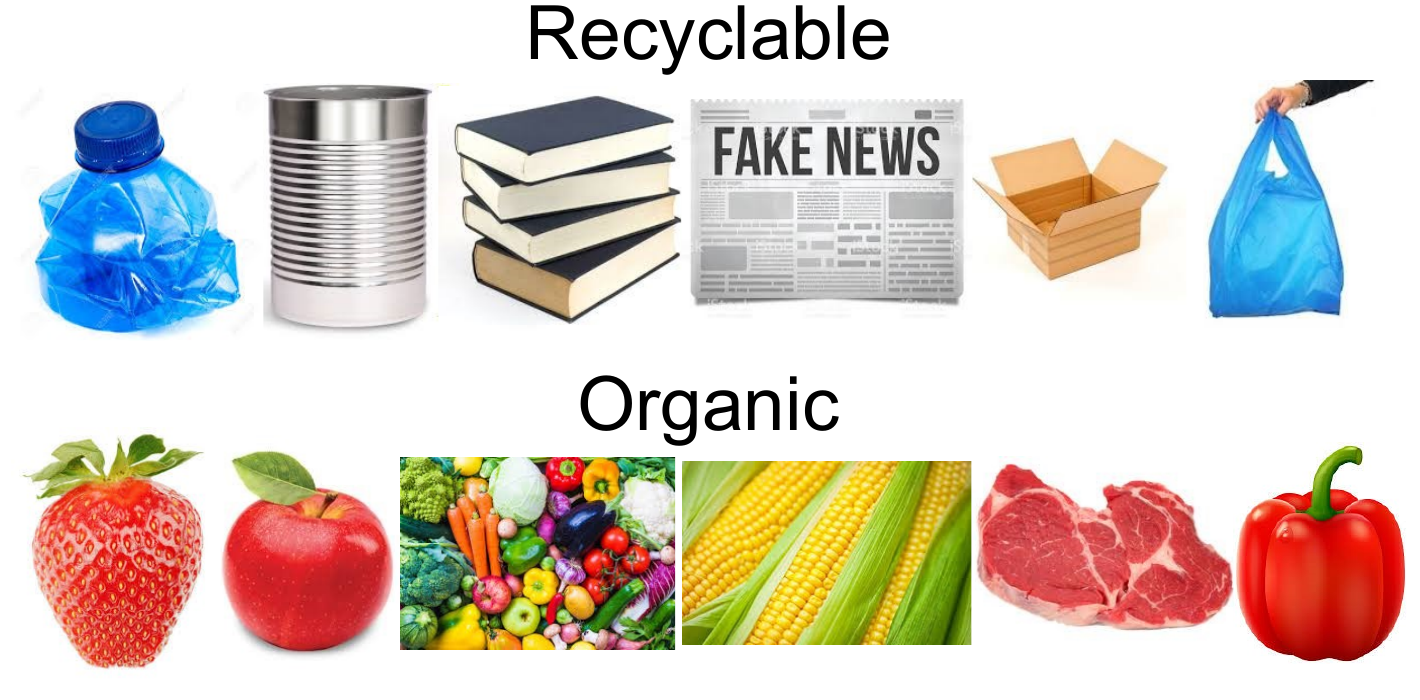} 
  \caption{Sample images of recyclable and organic materials used for CNN classification. The dataset contains 25,000 images, with 60\% used for training and 40\% for testing.}

  \label{Sample_images}
\end{figure}

\begin{table*}[t!]
\centering
\captionsetup[table]{font=tiny, skip=2pt} 
\caption{Rastrigin function convergence: Impact on mean $(\mu)$ PSO iterations and standard deviation $(\sigma)$ under varying exploration ($c_1$) and exploitation ($c_2$) settings.}

\resizebox{\textwidth}{!}{%
\begin{tabular}{|c|cc|cc|cc|}
\hline
\multirow{2}{*}{\textbf{No.   of Particles}} &
  \multicolumn{2}{c|}{\textbf{$c_1$ = $c_2$}} &
  \multicolumn{2}{c|}{\textbf{$c_1$ = 0.8, $c_2$ = 0.2}} &
  \multicolumn{2}{c|}{\textbf{$c_1$ = 0.2, $c_2$ = 0.8}} \\ \cline{2-7} 
 &
  \multicolumn{1}{c|}{\textbf{Mean$(\mu)$}} &
  \textbf{Standard Deviation$(\sigma)$} &
  \multicolumn{1}{c|}{\textbf{Mean$(\mu)$}} &
  \textbf{Standard Deviation$(\sigma)$} &
  \multicolumn{1}{c|}{\textbf{Mean$(\mu)$}} &
  \textbf{Standard Deviation$(\sigma)$} \\ \hline
\textbf{20}  & \multicolumn{1}{c|}{168.4} & 12.76 & \multicolumn{1}{c|}{172} & 15.97  & \multicolumn{1}{c|}{173.7} & 20.87 \\ \hline
\textbf{50}  & \multicolumn{1}{c|}{155.2} & 14.68 & \multicolumn{1}{c|}{155.1} & 14.49 & \multicolumn{1}{c|}{171.4} & 19.96 \\ \hline
\textbf{100} & \multicolumn{1}{c|}{146.5} & 10.2 & \multicolumn{1}{c|}{139.9} & 5.47 & \multicolumn{1}{c|}{154}   & 9.04 \\ \hline
\end{tabular}
}
\label{Exploration_Exploitation_variations}
\end{table*}


\begin{figure*}[h]
  \centering
  \includegraphics[scale=0.8]{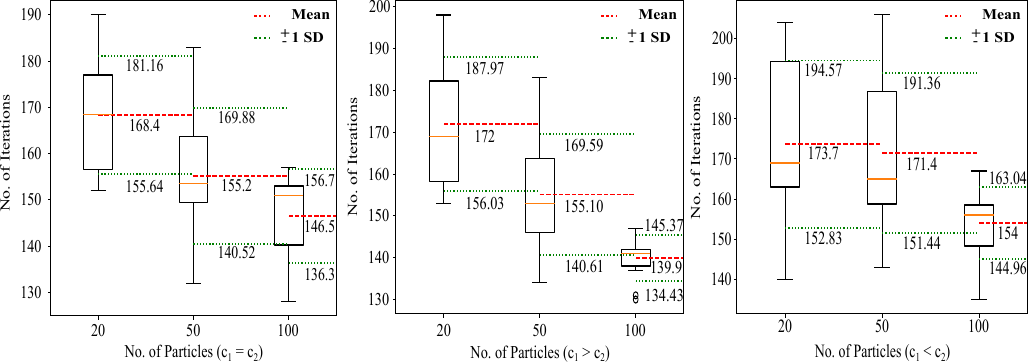} 
  \caption{Box plot showing the number of iterations (note: y-axis does not start at zero) required for Rastrigin function convergence using PSO with varying particle counts and exploration/exploitation settings. Each box represents 10 runs, with mean $(\mu)$ and standard deviation $(\sigma)$ of iteration counts.}

  \label{Rastrigin_PSO_only}
\end{figure*}


\subsection{\textbf{Classification Model}}

For our classification task, we used a dataset of 25,000 images labeled as (1) Organic or (2) Recyclable materials. Sample images are shown in Figure \ref{Sample_images}. The data was split into 60\% for training and 40\% for testing. A CNN was employed to perform the classification.

CNNs, inspired by the visual cortex, are feed-forward networks designed for automatic feature extraction and classification. Unlike traditional methods, they require minimal pre-processing, learning optimal filters during training. A typical CNN includes an input layer, convolutional hidden layers, and an output layer. Convolutional layers extract features through kernel-based operations, making CNNs effective for tasks such as image recognition, segmentation, and classification.

In our study, the CNN model classified random items as recyclable or organic, following the workflow in Figure \ref{methodology_diagram}, with the only change being the dataset and DL model.

To highlight optimization impact, we skipped image pre-processing and used raw images as CNN input. PSO initially optimized the CNN parameters, improving classification accuracy. LLM-driven PSO further refined these parameters, yielding even better results. This demonstrates the robustness and efficiency of our proposed optimization method.


\begin{table}[!h]
\centering
\captionsetup[table]{font=tiny, skip=2pt} 
\caption{LLM-driven PSO performance on the Rastrigin function: Llama3 converges faster than ChatGPT-3.5, with higher standard deviation in mean values.}

\small
\begin{tabular}{|c|cc|cc|}
\hline
\makecell[c]{\textbf{No. of}\\\textbf{Particles}} 
& \multicolumn{2}{c|}{\textbf{ChatGPT-3.5}} 
& \multicolumn{2}{c|}{\textbf{Llama3}} \\ \cline{2-5} 
 & \textbf{Mean$(\mu)$} & \textbf{Std. Dev.$(\sigma)$} 
 & \textbf{Mean$(\mu)$} & \textbf{Std. Dev.$(\sigma)$} \\ \hline
\textbf{20} & 173.9 & 7.82 & 154.7 & 12.73 \\ \hline
\textbf{50} & 148.6 & 13.01 & 142.0 & 18.95 \\ \hline
\textbf{100} & 144.3 & 13.6 & 142.2 & 15.13 \\ \hline
\end{tabular}
\label{Rastrigin_LLM_results}
\end{table}


\section{Results}
\label{sec:results}

We applied our proposed methodology to three distinct cases. Detailed explanations of the results and insights are provided in the following subsections.

\subsection{Rastrigin Function}

We evaluated the Rastrigin function, a standard PSO benchmark, to measure iterations required for convergence to the global minimum. Table \ref{Exploration_Exploitation_variations} presents results for PSO alone with varying particle counts (20, 50, 100) and configurations of exploration ($c_1$) and exploitation ($c_2$) parameters. 

For 20 particles, balanced exploration and exploitation ($c_1 = c_2$) achieved the lowest iterations with good stability. With 50 particles, both balanced ($c_1 = c_2$) and exploration-dominant ($c_1 = 0.8$, $c_2 = 0.2$) configurations performed similarly. For 100 particles, exploration-dominant settings provided faster and more stable convergence. Increasing particle count generally reduced iterations, as shown in Figure \ref{Rastrigin_PSO_only}.

\begin{figure*}
  \centering
  \includegraphics[scale=0.7]{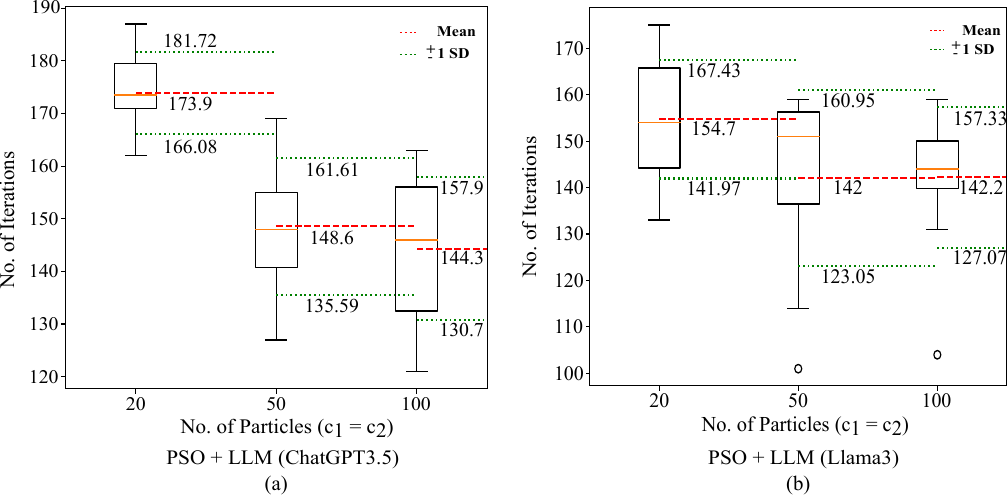} 
  \caption{Box plot of the number of iterations \textit{(y-axis does not start at zero)} required for Rastrigin function convergence using LLM-driven PSO with varying particle counts. Each box shows the mean $(\mu)$ and standard deviation $(\sigma)$ over 10 runs. (a) ChatGPT-3.5 results. (b) Llama3 results.}

  \label{Rastrigin_ChatGPT_Llama3}
\end{figure*}

Table \ref{Rastrigin_LLM_results} summarizes results from applying our LLM-augmented PSO scheme under similar particle settings. With exploration and exploitation balanced ($c_1 = c_2$), PSO iterated initially before LLMs (ChatGPT-3.5 or Llama3) provided recommendations. ChatGPT-3.5 showed no improvement with 20 particles but reduced iterations by 4.25\% and 1.50\% for 50 and 100 particles, respectively. Llama3 achieved 8.14\%, 8.50\%, and 2.94\% reductions for 20, 50, and 100 particles, respectively, though stability decreased for 50 particles. Results indicate faster convergence with our scheme compared to standalone PSO, as visualized in Figure \ref{Rastrigin_ChatGPT_Llama3} (a) for ChatGPT-3.5 and (b) for Llama3.

\begin{figure}[h]
  \centering
  \includegraphics[scale=.7]{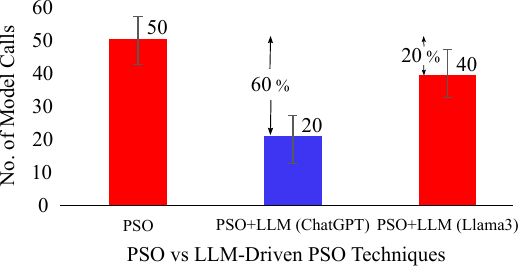} 
  \caption{Comparison of PSO and LLM-driven PSO in terms of model calls during regression, where an LSTM model is iteratively used to minimize the objective function for Air Quality Index forecasting.}

  \label{Regression_model_calls}
\end{figure}
\begin{figure}[h]
  \centering
  \includegraphics[scale=.7]{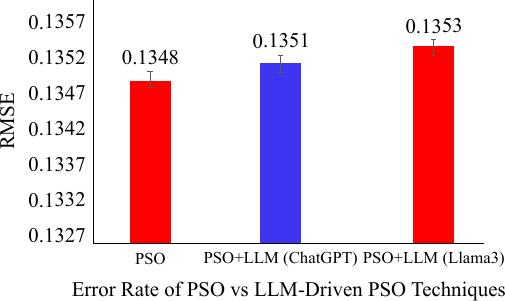} 
 \caption{Comparison of PSO and LLM-driven PSO (ChatGPT-3.5 and Llama3) in terms of error rate for the regression case \textit{(y-axis does not start at zero to better illustrate the data range)}. LLM-driven PSO achieves a cost value comparable to standard PSO.}

  \label{regression_barchart}
\end{figure}


\subsection{Regression Model}

We applied our methodology to optimize an LSTM model for AQI forecasting. First, we used the PSO algorithm to update the LSTM's layers and neurons dynamically. Then, we introduced our novel scheme, achieving the same cost value with fewer model calls by consulting LLMs (ChatGPT-3.5 and Llama3) after PSO iterations.

In our experiments, we tested various particle counts (5, 10, 15, 20), as shown in Table \ref{Best_number_of_particles}. Using 5 particles provided similar results to higher counts, making it the optimal choice for PSO in our scheme. We maintained balanced exploration and exploitation terms to optimize performance and reduce computational complexity.

\begin{table}[]
\centering
\caption{Global best cost across varying PSO particle counts shows minimal variation, supporting the use of five particles for improved efficiency.}
\resizebox{\columnwidth}{!}{%
\begin{tabular}{|c|c|c|c|c|}
\hline
\multicolumn{5}{|c|}{\textbf{Only PSO}} \\ \hline
\textbf{No. of Particles} & \textbf{5} & \textbf{10} & \textbf{15} & \textbf{20} \\ \hline
\textbf{Global Best Layer} & 2 & 2 & 2 & 2 \\ \hline
\textbf{Global Best Neuron} & 22 & 192 & 69 & 66 \\ \hline
\textbf{Global Best Cost} & \textbf{0.1343} & \textbf{0.1333} & \textbf{0.1336} & \textbf{0.1329} \\ \hline
\end{tabular}%
\label{Best_number_of_particles}
}
\end{table}

\begin{table*}[]
\centering
\caption{Regression results for an LSTM model optimized using three methods: (1) PSO alone, (2) LLM-driven PSO with ChatGPT-3.5, and (3) LLM-driven PSO with Llama3. A 95\% confidence interval was computed from three runs per approach. PSO-only used 10 iterations with 5 particles; ChatGPT-3.5 used 4 iterations, and Llama3 used 6–8 iterations—each with 5 particles—achieving comparable error rates.}

\begin{tabular}{|l|ccc|ccc|ccc|}
\hline
 & \multicolumn{3}{c|}{\textbf{Only PSO}} & \multicolumn{3}{c|}{\textbf{PSO + LLM (ChatGPT 3.5)}} & \multicolumn{3}{c|}{\textbf{PSO + LLM (Llama3)}} \\ \hline
\textbf{Layers} & \multicolumn{1}{c|}{2} & \multicolumn{1}{c|}{4} & 5 & \multicolumn{1}{c|}{3} & \multicolumn{1}{c|}{3} & 5 & \multicolumn{1}{c|}{3} & \multicolumn{1}{c|}{3} & 4 \\ \hline
\textbf{Neurons} & \multicolumn{1}{c|}{22} & \multicolumn{1}{c|}{200} & 200 & \multicolumn{1}{c|}{150} & \multicolumn{1}{c|}{105} & 200 & \multicolumn{1}{c|}{200} & \multicolumn{1}{c|}{186} & 154 \\ \hline
\textbf{RMSE} & \multicolumn{1}{c|}{0.1343} & \multicolumn{1}{c|}{0.1344} & 0.1358 & \multicolumn{1}{c|}{0.1343} & \multicolumn{1}{c|}{0.1355} & 0.1355 & \multicolumn{1}{c|}{0.1357} & \multicolumn{1}{c|}{0.1347} & 0.1355 \\ \hline
\textbf{Confidence Interval} & \multicolumn{3}{c|}{0.1327, 0.1369} & \multicolumn{3}{c|}{0.1333, 0.1369} & \multicolumn{3}{c|}{0.1347, 0.1359} \\ \hline
\textbf{PSO Iterations} & \multicolumn{3}{c|}{\textbf{10}} & \multicolumn{3}{c|}{\textbf{4}} & \multicolumn{1}{c|}{\textbf{8}} & \multicolumn{1}{c|}{\textbf{6}} & \textbf{6} \\ \hline
\end{tabular}%
\label{Regression_all_cases}
\end{table*}


For 10 iterations with 5 particles, PSO alone achieved the desired cost value after 50 model calls (Table \ref{Regression_all_cases}). With our approach, we ran PSO for 2 iterations, consulted the LLMs for recommendations, and continued iterating until the cost function reached the target. Using ChatGPT-3.5, we achieved the desired outcome with only 20 model calls (60\% improvement), while Llama3 required 30-40 calls (20\%-40\% improvement), as shown in Figure \ref{Regression_model_calls}. The 95\% confidence interval of our results included the PSO-only cost value, confirming the efficacy of our approach.

Figure \ref{regression_barchart} compares RMSE with a 95\% confidence interval for PSO and LLM-driven PSO methods, demonstrating the performance gains and reduced model calls with our approach. Llama3 required more iterations and model calls than ChatGPT-3.5 to achieve similar accuracy, indicating potential optimization opportunities for Llama3's prompt processing and architecture.



\subsection{Classification Model}

In our classification task, we used random images to distinguish between Recyclable and Organic materials, employing a CNN for optimization. Initially, we applied PSO with 5 particles and 10 iterations, resulting in a mean accuracy of 86.26\% with a 95\% confidence interval after 50 model calls.

\begin{figure}[h]
  \centering
  \includegraphics[scale=.7]{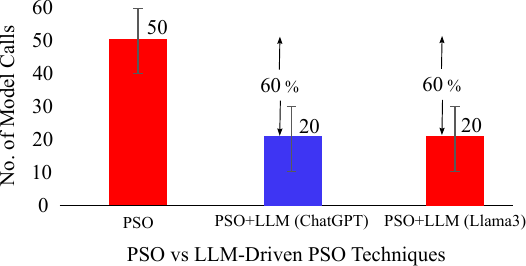} 
  \caption{Comparison of PSO and LLM-driven PSO Techniques in terms of Model Calls for the Classification Task using a CNN to differentiate between Recyclable and Organic materials.}
  \label{classification_model_calls}
\end{figure}

We then introduced our novel approach, utilizing 2 iterations of PSO followed by LLM suggestions (ChatGPT-3.5 or Llama3) to adjust particle positions and velocities. This method reduced the number of model calls to 20 while maintaining similar accuracy, achieving a 60\% reduction compared to the PSO-only approach as shown in Figure \ref{classification_model_calls}. These results are summarized in Table \ref{Classification_all_cases}, with a 95\% confidence interval for both ChatGPT and Llama3.

\begin{figure}[h]
  \centering
  \includegraphics[scale=.7]{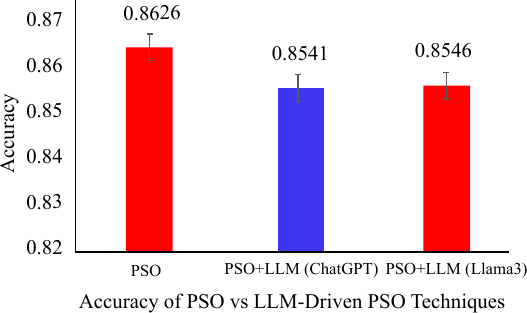} 
  \caption{Comparison of PSO and LLM-driven PSO (ChatGPT-3.5 and Llama3) in terms of accuracy for the classification case \textit{(y-axis does not start at zero to better illustrate the data range)}. The accuracy of LLM-driven PSO is comparable to that of standard PSO.}

  \label{classification_barchart}
\end{figure}

Figure \ref{classification_barchart} compares the accuracy of both PSO and LLM-driven PSO methods, showing almost identical accuracy with significantly fewer model calls during optimization.

\begin{table*}[]
\centering
\caption{Classification results for a CNN model optimized using three approaches: (1) PSO alone, (2) LLM-driven PSO with ChatGPT-3.5, and (3) LLM-driven PSO with Llama3. A 95\% confidence interval was computed from three runs per method. PSO-only used 10 iterations with 5 particles, while both LLM-driven approaches used 4 iterations with 5 particles, achieving comparable accuracy.}

\begin{tabular}{|l|ccc|ccc|ccc|}
\hline
 & \multicolumn{3}{c|}{\textbf{Only PSO}} & \multicolumn{3}{c|}{\textbf{PSO + LLM (ChatGPT 3.5)}} & \multicolumn{3}{c|}{\textbf{PSO + LLM (Llama3)}} \\ \hline
\textbf{Layers} & \multicolumn{1}{c|}{4} & \multicolumn{1}{c|}{5} & 3 & \multicolumn{1}{c|}{4} & \multicolumn{1}{c|}{3} & 4 & \multicolumn{1}{c|}{3} & \multicolumn{1}{c|}{3} & 3 \\ \hline
\textbf{Filters} & \multicolumn{1}{c|}{16} & \multicolumn{1}{c|}{21} & 24 & \multicolumn{1}{c|}{16} & \multicolumn{1}{c|}{20} & 58 & \multicolumn{1}{c|}{32} & \multicolumn{1}{c|}{37} & 21 \\ \hline
\textbf{Accuracy} & \multicolumn{1}{c|}{0.8622} & \multicolumn{1}{c|}{0.8638} & 0.8619 & \multicolumn{1}{c|}{0.8515} & \multicolumn{1}{c|}{0.8587} & 0.8521 & \multicolumn{1}{c|}{0.8553} & \multicolumn{1}{c|}{0.8485} & 0.8601 \\ \hline
\textbf{Confidence Interval} & \multicolumn{3}{c|}{0.8601, 0.8652} & \multicolumn{3}{c|}{0.8442, 0.8640} & \multicolumn{3}{c|}{0.8480, 0.8612} \\ \hline
\textbf{PSO Iterations} & \multicolumn{3}{c|}{\textbf{10}} & \multicolumn{3}{c|}{\textbf{4}} & \multicolumn{3}{c|}{\textbf{4}} \\ \hline
\end{tabular}%
\label{Classification_all_cases}
\end{table*}


\begin{table*}[]
\centering
\caption{Number of PSO iterations before LLM use with varying particle counts, while keeping exploration and exploitation parameters constant.}
\begin{tabular}{|c|c|c|c|c|c|}
\hline
\multicolumn{6}{|c|}{\textbf{Exploration = 0.5, Exploitation = 0.5}} \\ \hline
\multirow{2}{*}{\textbf{No. of Particles}} & \multicolumn{5}{c|}{\textbf{LLM (ChatGPT-3.5)-driven PSO}} \\ \cline{2-6} 
 & \textbf{10 iterations} & \textbf{20 iterations} & \textbf{30 iterations} & \textbf{40 iterations} & \textbf{50 iterations} \\ \hline
\textbf{20}  & 123 & 140 & 147 & 168 & 165 \\ \hline
\textbf{50}  & 132 & 141 & 154 & 165 & 177 \\ \hline
\textbf{100} & 163 & 169 & 172 & 168 & 173 \\ \hline
\end{tabular}
\label{Rastrigin_function_table}
\end{table*}


\section{Lessons Learned}
\label{sec:lessons_learned}

This paper introduces a novel methodology for optimizing DL model hyper-parameters by integrating LLMs with the PSO algorithm, demonstrating enhanced efficiency and earlier convergence across various scenarios. Based on our results, we summarize our findings and lessons learned as follows.
\begin{itemize}
    \item In the Rastrigin function case, we varied exploration and exploitation parameters. With fewer particles, balanced settings performed best; with more particles, exploration-dominant configurations offered greater stability.

    \item In our proposed approach, where PSO runs for a few iterations before consulting LLMs for optimal particle positions, we observed that reducing the initial PSO iterations at each step resulted in earlier convergence, as shown in Table \ref{Rastrigin_function_table} for the Rastrigin function case.    
    \item For DL models like LSTMs and CNNs, five particles were sufficient to reach optimal cost. Higher particle counts increased complexity without improving results. Our method consistently converged within four iterations, confirming that fewer PSO steps can lead to earlier convergence.

    \item A key takeaway from our research is the potential to extend our approach to other metaheuristic techniques. By integrating LLMs with various optimization methods, we can reduce computational costs and accelerate model convergence by decreasing the number of iterations needed. In Genetic Algorithms, LLMs can enhance the selection process by guiding parent selection for the next generation and can also be applied in the mutation phase to introduce random variations, maintaining genetic diversity and preventing premature convergence.
    \item Based on our observations, Llama3 outperformed ChatGPT-3.5 with a slightly higher standard deviation $(\sigma)$ in the Rastrigin function case. In the regression model, ChatGPT demonstrated greater consistency and produced better results compared to Llama3. In the classification case, both LLMs performed similarly well.
   \item Integrating LLMs into PSO significantly improves convergence and reduces computational cost by 20\% to 60\% compared to traditional PSO methods~\cite{qolomany2017parameters, fouad2021hyper}, without sacrificing accuracy.

    \item The reduced computational cost of our approach makes it well-suited for resource-constrained settings such as IoT, edge computing, and real-time analytics.

    \item The effectiveness of our LLM-driven PSO depends on prompt quality, highlighting the need for automated prompt engineering. Future work includes testing on larger datasets and exploring complex tasks like multi-objective and dynamic optimization.

\end{itemize}

\section{Conclusion \& Future Work}
\label{sec:conclusion_and_future_work}

We propose and evaluate a novel approach to optimize deep learning models using Particle Swarm Optimization enhanced by Large Language Models. Our method was applied to three scenarios: optimizing the Rastrigin function, tuning an LSTM model for time series prediction, and optimizing a CNN model for image classification.

For the Rastrigin function, our PSO-LLM approach achieved faster convergence to the global minimum, reducing the number of iterations compared to traditional PSO alone. In LSTM-based time series prediction, we demonstrated that the PSO-LLM strategy delivered similar predictive results with fewer model evaluations, improving real-time decision-making efficiency. For image classification using CNNs, our method achieved comparable accuracy to traditional PSO while enhancing hyperparameter optimization with LLM insights.

Overall, our study contributes to DL optimization by combining PSO with LLMs (ChatGPT-3.5 and Llama3), improving convergence speed and optimization outcomes. Future work could explore extending this approach to other algorithms and evaluating its scalability with larger datasets and different LLMs. In future work, we plan to further validate the generalizability and robustness of our proposed LLM-enhanced PSO approach by conducting additional experiments using other standard benchmark optimization functions such as Sphere, Rosenbrock, and Ackley. These benchmarks will provide broader validation and help demonstrate the applicability of our method across diverse and complex optimization landscapes.

\bibliographystyle{IEEEtran}

\bibliography{sample.bib}

\begin{IEEEbiography}
[{\includegraphics[width=1.1in,height=1.4in,clip,keepaspectratio]{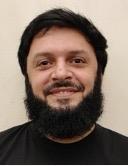}}]{Saad Hameed }
received his B.Sc. in Electrical Engineering from UET Peshawar, Pakistan, in 2010, and his M.S. in Electrical Engineering (Digital Systems and Signal Processing) from NUST, Pakistan, in 2015. He worked as a Research Assistant at CISNR and the Smart Machines and Robotics Technology Lab from 2011 to 2015. From 2016 to 2022, he served as a Metering Engineer at SNGPL, a major gas utility in Pakistan. Since 2022, he has been pursuing a Ph.D. in Computer Science at Hamad Bin Khalifa University, Qatar. His research focuses on optimizing deep learning models using Large Language Models (LLMs) to improve the efficiency of edge devices in smart city applications.

\end{IEEEbiography}
\begin{IEEEbiography}[{\includegraphics[width=1.1in,height=1.4in,clip,keepaspectratio]{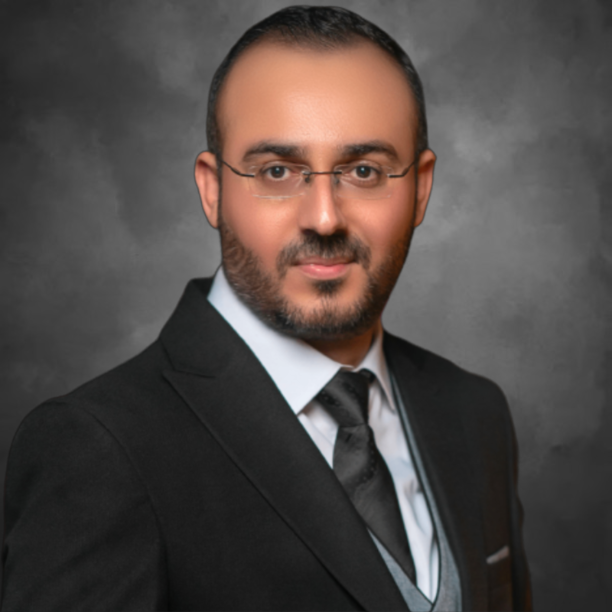}}]{Basheer Qolomany }
 received his Ph.D. in Computer Science from Western Michigan University, USA, in 2018. He is currently an Assistant Professor of AI and Computational Medicine at Howard University College of Medicine. His interdisciplinary research spans network science, evolutionary computation, and AI, with applications in computational medicine, population health, cybersecurity, and smart services. He applies techniques including NLP, deep learning, graph learning, swarm intelligence, and big data analytics. Dr. Qolomany has held academic and research roles at institutions including the University of Cincinnati, University of Nebraska, Kennesaw State University, Western Michigan University and the University of Duhok.
\end{IEEEbiography}
\begin{IEEEbiography}
[{\includegraphics[width=1.1in,height=1.4in,clip,keepaspectratio]{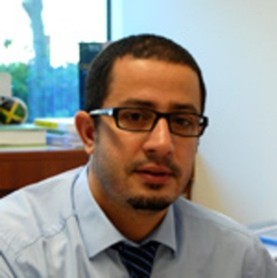}}]{Samir Brahim Belhaouari}
 received his Master’s in Telecommunications and Network from the Institut National Polytechnique of Toulouse, France, in 2000, and his Ph.D. in Mathematics from Federal Polytechnic School of Lausanne, Switzerland, in 2006. He is currently an Associate Professor in the Division of Information and Communication Technologies at Hamad Bin Khalifa University, Qatar. He has held research and teaching positions at institutions across Russia, Saudi Arabia, the UAE, Malaysia, and Switzerland. His research interests include applied mathematics, statistics, data analysis, AI, and image and signal processing, with applications in biomedicine, bioinformatics, and forecasting.
\end{IEEEbiography}

\begin{IEEEbiography}
 [{\includegraphics[width=1.1in,height=1.4in,clip,keepaspectratio]{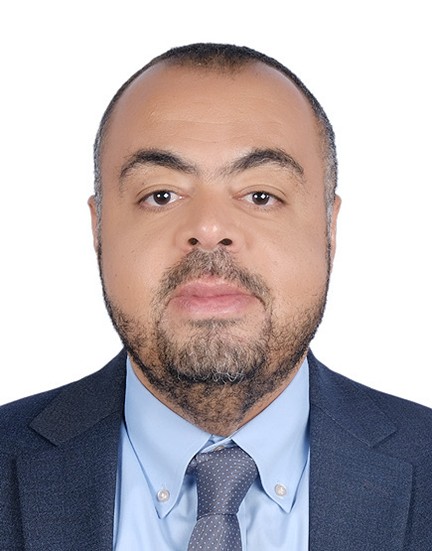}}]{Mohamed Abdallah }
  is a Professor and Associate Dean of Undergraduate Studies and Quality Assurance at the College of Science and Engineering, Hamad Bin Khalifa University (HBKU), Qatar. He received his M.Sc. and Ph.D. from the University of Maryland, College Park, in 2001 and 2006, respectively. His research focuses on AI for communications, including 6G networks, wireless security, electric vehicles, and smart grids. He has published over 220 papers, contributed to four book chapters, and holds four patents. His recognitions include the Research Fellow Excellence Award at Texas A\&M University at Qatar (2016) and multiple IEEE Best Paper Awards. He was also a Nortel Networks Industrial Fellow (1999–2003).

\end{IEEEbiography}

\begin{IEEEbiography}
[{\includegraphics[width=1.1in,height=1.4in,clip,keepaspectratio]{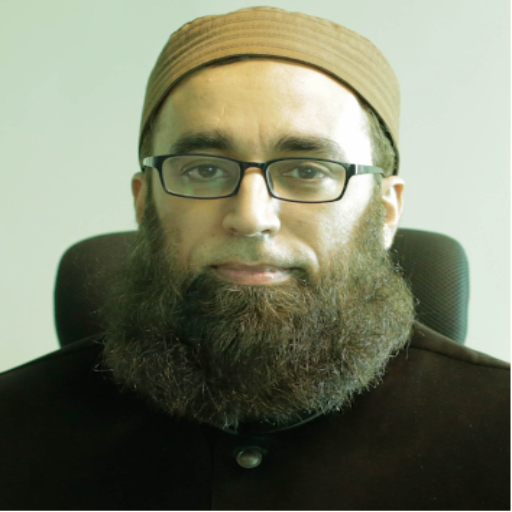}}]{Junaid Qadir }
  is a Professor of Computer Engineering at Qatar University, where he directs the IHSAN Research Lab. His research spans computer systems, networking, applied machine learning, ICT for development (ICT4D), human-centric AI, and the ethics of technology and data science. He has published over 150 peer-reviewed articles in top-tier venues such as IEEE JSAC, CST, TMC, and Communications Magazine. He received Pakistan’s Higher Education Commission Best University Teacher Award (2012–2013) and has secured research funding from Facebook, QNRF, and HEC Pakistan. Dr. Qadir is an ACM Distinguished Speaker (2020–2023) and a Senior Member of IEEE and ACM.
\end{IEEEbiography}

\begin{IEEEbiography}
[{\includegraphics[width=1.1in,height=1.4in,clip,keepaspectratio]{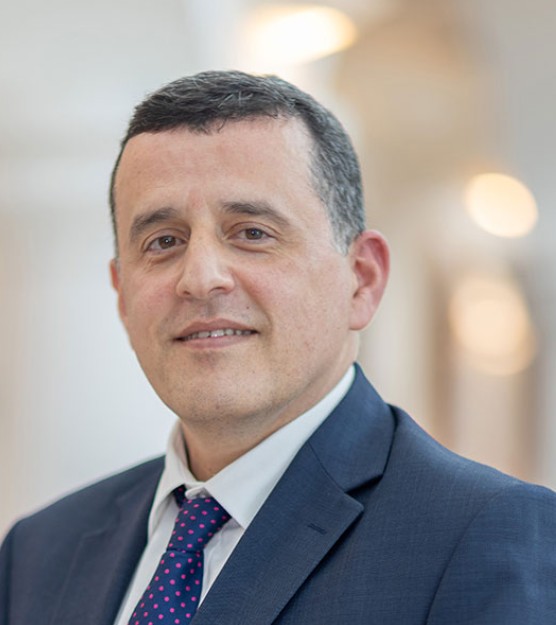}}]{Ala Al-Fuqaha }
  is currently serving as Acting Provost and Associate Provost for Teaching and Learning at Hamad Bin Khalifa University (HBKU). He joined HBKU in 2018 as a professor in the Information and Computing Technology (ICT) division of the College of Science and Engineering (CSE). Before HBKU, he was a professor and the director of the NEST research lab in the computer science department of Western Michigan University. In 2006, Dr. Al-Fuqaha was awarded the Outstanding New Educator Award from the College of Engineering and Applied Sciences at Western Michigan University. In 2014, he received the Outstanding Researcher Award from the College of Engineering and Applied Sciences at Western Michigan University. In 2018, he received the Best Survey Award from the IEEE Communications Society.
\end{IEEEbiography}
\vfill\pagebreak

\end{document}